\documentclass[11pt]{article}
\usepackage{graphicx,amsmath,amsfonts,amscd,amssymb,bm,cite,epsfig,epsf,url,color,algorithm,algorithmic,multirow}
\usepackage{appendix}
\usepackage{booktabs}
\usepackage{fullpage} \usepackage[small,bf]{caption}
\usepackage{subfigure}
\numberwithin{equation}{section}

\graphicspath{ {./} }
\title{Conditionally Learn to Pay Attention for Sequential Visual Task}

\author{Jun~He$^{\dagger}$, Quan-Jie Cao$^{\dagger}$, and Lei~Zhang$^{\ddagger}$\\
 \vspace{-.1cm}\\        
 $^{\dagger}$ School of Artificial Intelligence, \\
 Nanjing University of Information  Science and Technology, Nanjing, China\\ 
jhe@nuist.edu.cn, caodada1118@163.com \\
  \vspace{-.3cm}\\
  $^{\ddagger}$ School of Electrical and Automation Engineering, \\
  Nanjing Normal University, Nanjing, China\\
leizhang@njnu.edu.cn \\
  \vspace{-.3cm}\\
}
        
\date{November 9, 2019}

\begin{document}
%\thanks{Jun He is with the College of Electronic and Information Engineering, 
%Nanjing University of Information Science and Technology, Nanjing,
%Jiangsu, China (hejun.zz@gmail.com).}% <-this % stops a space
%\thanks{Laura Balzano is with the Department of Electrical and Computer Engineering, University of Wisconsin-Madison,
%Wisconsin, U.S., (sunbeam@ece.wisc.edu).}% <-this % stops a space
%\thanks{John C.S. Lui is with the Department of Computer Science and Engineering, the Chinese University of Hong Kong,
%Hong Kong, (cslui@cse.cuhk.edu.hk).}% <-this % stops a space
% make the title area
\maketitle 

\begin{abstract}
Sequential visual task usually requires to pay attention to its current interested object conditional on its previous observations. Different from popular soft attention mechanism, we propose a new attention framework by introducing a novel \emph{conditional global feature} which represents the weak feature descriptor
of the current focused object. Specifically, for a standard CNN (Convolutional Neural Network) pipeline, the convolutional layers with different
receptive fields are used to produce the attention maps by measuring how the convolutional features align to the \emph{conditional global feature}. The \emph{conditional global feature} can be generated by different recurrent structure according to different visual tasks, such as a simple recurrent neural network for multiple objects recognition, or a moderate complex language model for image caption. Experiments show that our proposed conditional attention model achieves the best performance on the SVHN (Street View House Numbers) dataset with / without extra bounding box; and for image caption, our attention model generates better scores than the popular soft attention model.
%\boldmath
\end{abstract}


%\IEEEpeerreviewmaketitle

\section{Introduction}
\label{sec:1}
Recent successes in machine translation~\cite{bahdanau2014neural}, speech recognition~\cite{graves2013speech}, and image caption~\cite{vinyals2015show} have witnessed
the important role of attention mechanism.
In computer vision, like human visual system, attention does not need to focus on
the whole image, but only on the salient areas of the image. %For example, attention mechanism plays a significant role in
For example, ~\cite{xu2015show},~\cite{you2016image},~\cite{chen2017sca},~\cite{anderson2018bottom} embedded attention mechanism into image caption which enables the model to learn to automatically generate a caption describing the content of an image. Subsequently, attention approaches were introduced into the emerging visual question answering task (VQA) which greatly improved the overall performance~\cite{xu2016ask}~\cite{lu2016hierarchical}~\cite{anderson2018bottom}.

Recently, ~\cite{jetley2018learn}
proposed a novel end-to-end trainable attention module for convolutional neural network architectures.
The core idea of their work lies in estimating the attention maps by
measuring how the local convolutional feature aligns to the global feature, which is different from
the previous attention approaches.
This novel attention approach experimentally proves its capability on weakly supervised object recognition / query though, it is only suitable for one object, saying that the trained global feature only represents a specific object according to the image label.

\begin{figure*}[htb]
\centering
\includegraphics[width=1.0\textwidth]{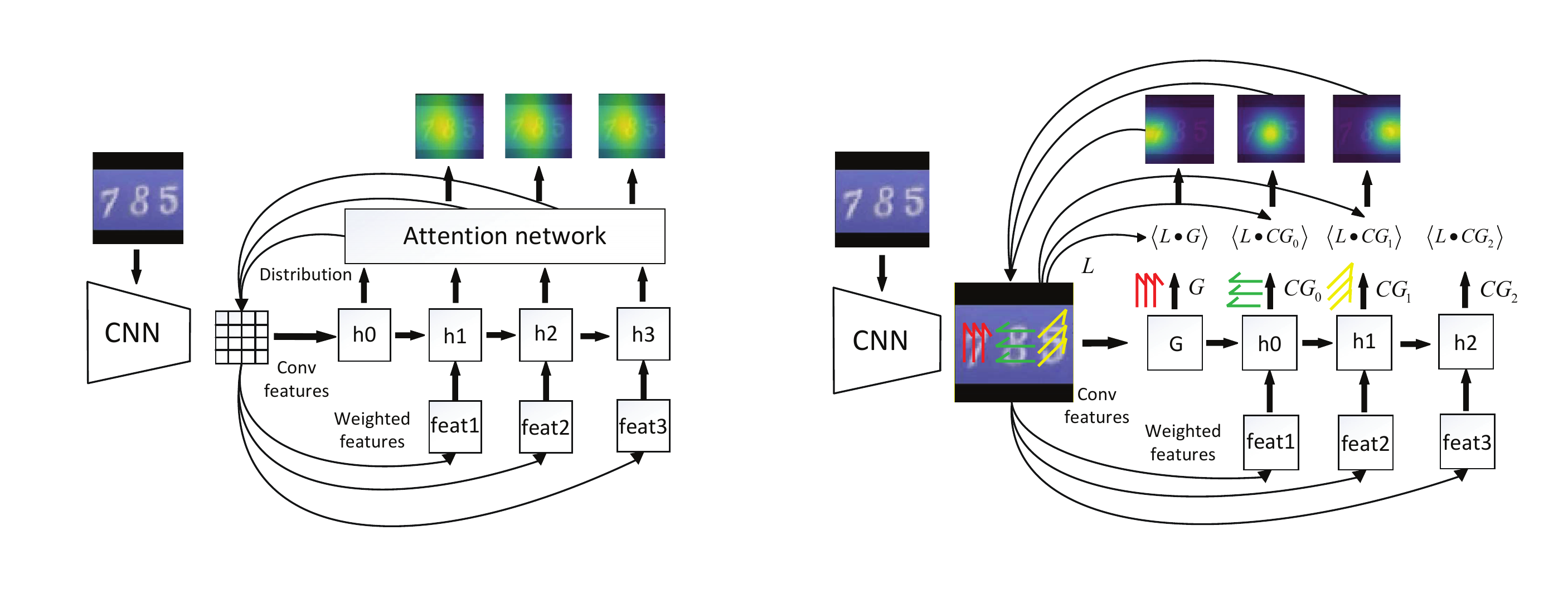}
\caption{Illustration of the difference between the popular soft attention architecture~\cite{xu2015show}(\textbf{left}) and our proposed conditional attention framework(\textbf{right}).
In our framework, '\emph{G}' denotes the global feature, '\emph{CG}' denotes the {\it conditional global feature}, and '\emph{L}' denotes convolutional features.}
\label{fig:1}
\end{figure*}

In this paper, we take a further step on the work of ~\cite{jetley2018learn} to extend its capability of sequential visual tasks, such as multiple objects recognition and image caption. Our work is inspired by the recent works on employing attention in image caption ~\cite{xu2015show}~\cite{you2016image}~\cite{chen2017sca}~\cite{anderson2018bottom}
and the new attention mechanism in ~\cite{jetley2018learn}, in which we propose a novel conditional attention framework
for the sequential visual tasks.
%, \emph{i.e.} multiple object recognition and image caption.
In order to
generate the attention map sequentially for each object, we design a {\it conditional global feature} and represent
it as a feature descriptor of the current focused object. Then the attention feature vector is sequentially produced through a compatibility function between convolutional features and \emph{conditional global feature} in our framework.
%The optimization procedure is end-to-end, it allows the model to be trained directly with respect to a given task,
%and uses backpropagation to train the neural network components.
Note that our new conditional attention framework is different from the popular soft attention architecture proposed by~\cite{xu2015show}.  Instead of predicting the probability distribution between image pixels through an attention network, we use dot product as a compatible function between local features and {\it conditional global feature} to get a map score that highlights the relevant areas of the image and suppresses the background information. 
Fig.\ref{fig:1} shows the difference between the popular soft attention model and our proposed model.

The contribution of our work is threefold.
	Firstly, unlike the popular soft attention model~\cite{xu2015show}, a very different conditional attention framework is proposed to tackle sequential visual tasks.
	Secondly, we introduce the novel  {\it conditional global feature} which can be regarded as a weak feature descriptor for a particular object in a sequential visual task. Then the attention map for each object can be simply estimated by measuring how the local convolutional feature aligns to the {\it conditional global feature}.
	Thirdly, for the popular multi-digit ``Street View House Number'' (SVHN) dataset, our new attention model achieves the  best performance over other state of the arts with / without extra bounding box.

\section{Related Work}
\label{sec:2}

Attention can be achieved in two ways, post-hoc mechanism and trainable attention in CNNs. ~\cite{simonyan2013deep} computes a class map to capture internal changes of deep convolutional  neural network for image
classification. Subsequently, ~\cite{cao2015look} proposed feedback CNNs to produce a saliency map to
show its attention on expected objects in the image. Notably, all the methods 
produce attention maps by passing through the well trained CNN models, which are called post-hoc processing.

In recent years, many studies demonstrated that the methods of obtaining attention maps by optimizing the weight of attention modules in the process of training CNNs can achieve better performance.
Trainable attention in CNNs are divided into two categories: hard attention and soft attention.
For the former, it is a stochastic sampling method and thus is non-differential, in which the attention module must be trained via Reinforcement Learning~\cite{sutton2000policy} methods, for example the work of recurrent attention model (RAM) ~\cite{mnih2014recurrent}.
For the latter, soft attention computes the weight vector as the attention probability, which is differentiable and can be easily trained.
Trainable attention in CNNs especially soft attention has been applied in a variety of visual tasks.
For example, attention can be applied to query-based tasks \cite{nema2017diversity},~\cite{xu2015show},~\cite{song2017unified},~\cite{xu2016ask},
and visual question answering tasks (VQA)~\cite{xu2016ask},~\cite{lu2016hierarchical},~\cite{anderson2018bottom}.
%What they have in common is that they use a high-dimensional vector to
%represent the input image, and guide the model to pay attention to the relevant parts of the images.
Especially, for query-based task, ~\cite{jetley2018learn} introduced a novel learn to pay attention method
 which directly uses a learned global feature to query images different from previous methods performing query by one-hot encoding of category labels~\cite{seo2016progressive}.
 However, the learned global feature does not consider its variability conditional on its context information, which prevents this new attention method from sequential visual tasks, such as weakly supervised multiple objects recognition task and image caption.
% However, since this new attention method depends on the learned global feature which does not consider its variability conditional on its context information, which prevent  constrains its capability on sequential visual tasks.
% aims at single object tasks, and has no sequential conditional attention mechanism which needs to focus attention on different objects of images, $i.e.$ weakly multiple objects recognition task and image caption.

%Inspired by the successful use of sequence to sequence training with neural networks in machine translation~\cite{cho2014learning},~\cite{bahdanau2014neural}
%and application of attention mechanism in visual tasks,
%Learning phrase representations using RNN encoder-decoder for statistical machine translation
%Neural machine translation by jointly learning to align and translate}

Recent image caption methods embedded attention into encoder-decoder framework~\cite{xu2015show},~\cite{you2016image},~\cite{chen2017sca},~\cite{anderson2018bottom}.
For example, two popular attention methods (soft and hard) were proposed in \cite{xu2015show} which not only can generate meaningful words  but can highlight the corresponding region of interest in images.
To enhance image caption, \cite{you2016image} combines both top-down and bottom-up approaches to fuse the extracted features of both sides.
%However, these spatial attentions applied to the output of a CNN is not enough to represent region features in images.
\cite{chen2017sca} incorporates spatial and channel-wise attentions in a convolutional neural network and
these attentions are embedded into different layers to represent multi-scale  features.
In order to highlight the image regions more accurately,~\cite{anderson2018bottom} utilizes bottom-up attention to obtain salient image regions
by Faster R-CNN like technique~\cite{ren2015faster},
and combines top-down attention as a language model to produce more coherent sentences.

\section{Conditional Attention Model}
\label{sec:3}

In this section, we first introduce the {\it conditional global feature}, the key idea of our proposed conditional attention framework for sequential visual task.
For multiple objects recognition, a simple recurrent neural network can be used for generating the {\it conditional global feature}. % for multiple objects recognition the first conditional attention model is described.
%which accepts the images containing multiple objects as inputs and output the corresponding category labels.
For image captioning, we demonstrate that a language model can be easily incorporated into this attention framework.
% which is an advanced version namely caption model.	

\subsection{Conditional Global Feature}

Our conditional attention framework is mainly inspired by~\cite{jetley2018learn} which
introduces a compatibility measure between local features and global features.
For sequential visual tasks, processing objects of an image usually consists of \emph{T} steps.
At each step \emph{t}, the model needs to produce an attention map of the current object and its corresponding attention feature.
Then the trained global feature must change accordingly to represent different objects, so the recurrent structure can be naturally exploited
to provide the conditional information for the variable global feature. Thus we propose the \emph{conditional global feature}, output of the recurrent structure given the previous attention feature $g_{t-1}$ and the last recurrent state $h_{t-1}$. In this paper, the \emph{conditional global feature} refers to the output vector of each LSTM cell~\cite{hochreiter1997long}:
% as discussed shortly, the trained global features must represent the feature information of different objects
% according to the change of focus object at each step, subsequently, we propose a \emph{conditional global feature} as a weak feature descriptor of the different focused object.
% In this paper, the \emph{conditional global feature} refers to the output vector of each LSTM cell:
\begin{equation} \label{eq:CGt}
C{G_t} = LSTM\left( {{g_{t - 1}},{h_{t - 1}}} \right)
\end{equation}
Note that, $CG_t$ is actually the updated recurrent state of LSTM $h_t$,  but it can be regarded as the weak feature descriptor of the focused object, which is the key design of our conditional attention framework.
%where $g_{t-1}$ is the previous attention feature and $h_{t-1}$ is the previous hidden state of LSTM~\cite{hochreiter1997long}.

\begin{figure}[htb]
\centering
\includegraphics[width=.5\textwidth]{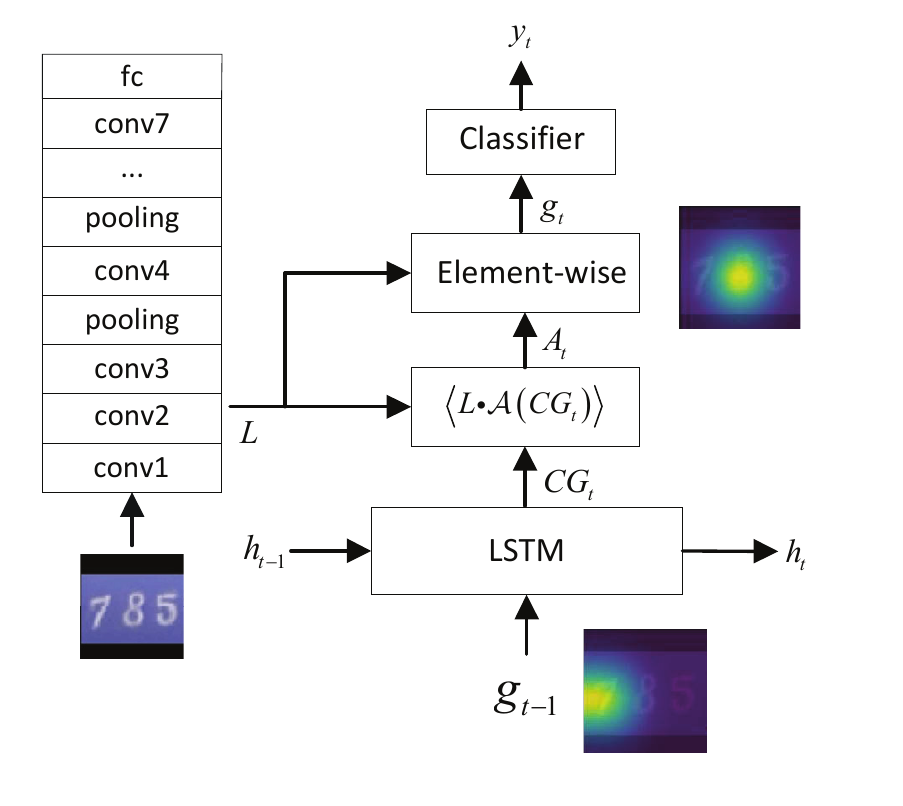}
\caption{Overview of the conditional attention model for multiple objects recognition, `$C{G_t}$' denotes our {\it conditional global feature} and `$L$' denotes CNN features. The detailed structure of convolution neural network is listed in the Appendix.}
\label{fig:cg}
\end{figure}

For multiple objects recognition task, the conditional attention model combined with a simple recurrent network is shown in Fig.~\ref{fig:cg}.
Once the \emph{conditional global feature} $C{G_t}$ is calculated, the attention maps of the current focused object can be generated by the conditional attention submodule, referring to Section~\ref{subsec:3.2} for details. Notably, at the first step, the model
directly estimates the attention map of the first object through compatibility function between local CNN features and the original global feature outputed by the last {\it fc} layer.

\subsection{Conditional Attention Submodule}
\label{subsec:3.2}

This section describes the details of attention generation, saying that how \emph{conditional global feature} aligns to local CNN features.
At each step \emph{t}, we denote ${L^s} = \{ l_1^s,l_2^s, \cdot  \cdot  \cdot ,l_n^s\} $ as the set of local feature vectors extracted at
a given intermediate layer $s \in \{ 1, \cdot  \cdot  \cdot ,S\} $ in the CNN pipeline, where $l_i^s$ is the \emph{i-th} feature
vector of \emph{n} total spatial locations in the local feature of layer \emph{s}.
Since the \emph{conditional global feature} vector $C{G_t}$ may have different dimension with  $l_i^s$, a linear mapping $\mathcal{A}$ is applied to $C{G_t}$ to keep the dimension consistent.
Then, dot product is used to measure the compatibility between $C{G_t}$  and $l_i^s$ as shown in Equation~\eqref{eq:compatibility}:

\begin{equation}\label{eq:compatibility}
c_i^s = \left\langle {l_i^s,\mathcal{A}(C{G_t})} \right\rangle ,i \in \left\{ {1 \cdot  \cdot  \cdot n} \right\}
\end{equation}
%On the premise that compatibility function requires two vectors of equal dimension,
%we use a linear mapping from $h_t$ to the dimensionality of $l_i^s$ to
%limit the parameters at the classification stage.
By calculating the compatibility function, we can get a series of compatibility
scores $C = \{ c_1^s,c_2^s, \cdot  \cdot  \cdot ,c_n^s\}$, which are then normalized by a softmax operation
\begin{equation}
a_i^s = \frac{{\exp \left( {c_i^s} \right)}}{{\sum\nolimits_j^n {\exp \left( {c_j^s} \right)} }},i \in \left\{ {1 \cdot  \cdot  \cdot n} \right\}.
\end{equation}

The normalized compatibility scores ${A^s} = \left\{ {a_1^s,a_2^s, \cdot  \cdot  \cdot a_n^s} \right\}$ are then used to produce
vector $g_t^s$ for layer \emph{s} by element-wise weighted averaging the local features:
\begin{equation}
g_t^s = \sum\nolimits_{i = 1}^n {a_i^s}  \cdot l_i^s
\end{equation}
%The obtained vector $g_t^s$ is a weighted combination feature of layer \emph{s} at step \emph{t},
To produce multiple scale attentions, %considering that the layer $\left( {S > 1} \right)$ in the CNN pipeline,
we can concatenate $g_t^s$ for different layers as $g_t = \left[ {{g_t^1},{g_t^2}, \cdot  \cdot  \cdot ,g_t^s} \right]$
which can provide more discriminative and complementary representation for a particular object at different scales.
%more abundantly express the complementary focus on different parts in the image at different scales.
Finally, we compute the conditional distribution over possible output through a classifier:
\begin{equation}
p\left( {{y_t}|{g_{t-1},{h_{t-1}}}} \right) = f\left( {g_t} \right)
\end{equation}
where \emph{f} is a classifier with a Softmax function that outputs the probability of $y_t$, and $g_t$ is the concatenated attention feature vector at time \emph{t}.

\subsection{Attention Model for Image Caption}
\label{subsec:3.3}

The above conditional attention model for multiple objects recognition is a basic model for sequential visual tasks.
%For our second version of conditional
However, for image captioning, unlike multiple object recognition, the model not only needs to output a word at each step,
but also needs a strong context dependence between words to produce a better sentence.
Here we show that our new conditional attention framework can incorporate a popular language model to solve image caption task.
%which called Caption Model in the sequel.

\begin{figure}[htb]
\centering
\includegraphics[width=.6\textwidth]{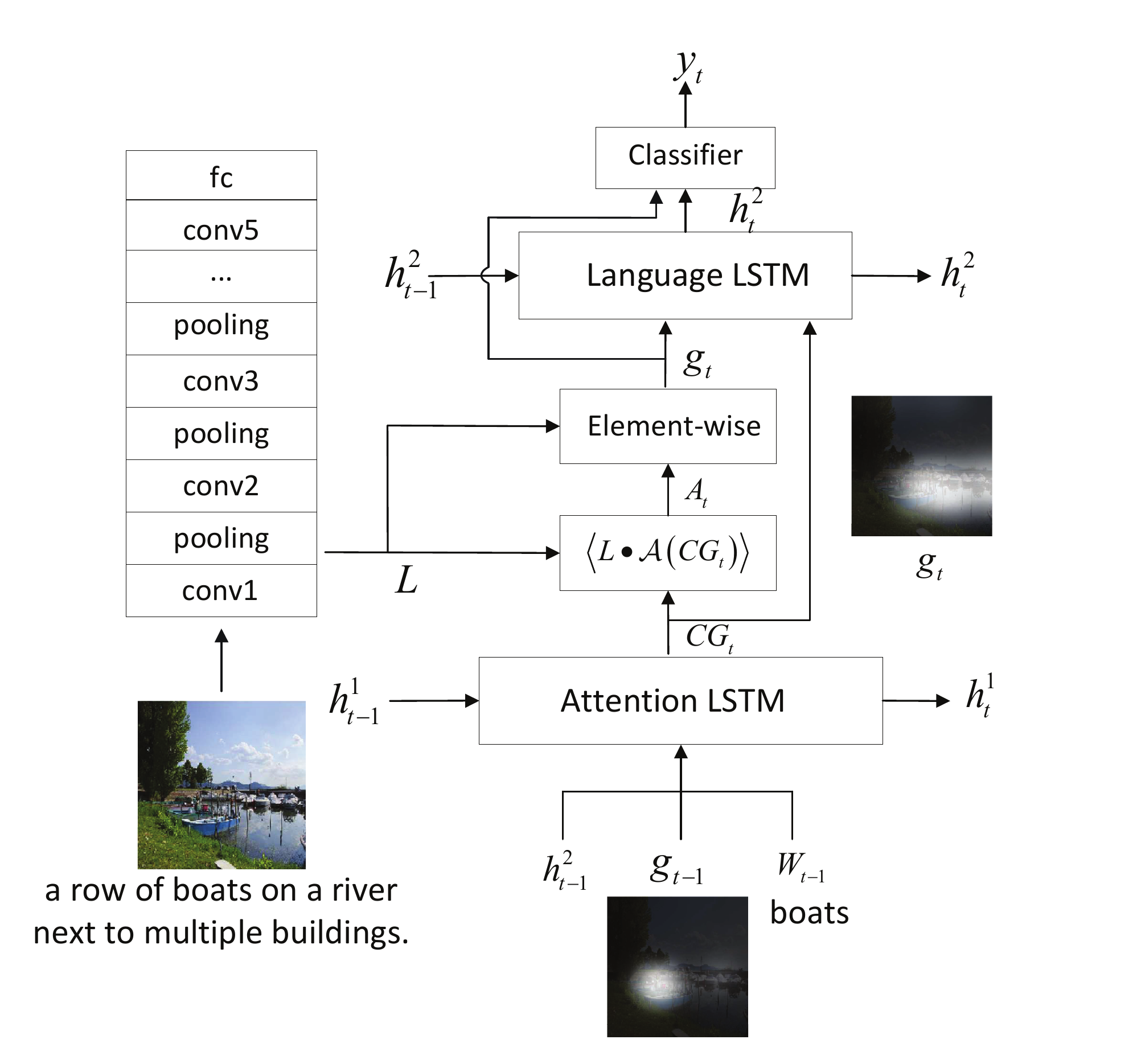}
\caption{Overall structure of the attention model for image captioning.
Two separated LSTM layers are used to generate visual attention features and context information for language model.
$C{G_t}$  not only represents the weak visual feature descriptor but provides the context visual information for a language model.}
\label{fig:cg2}
\end{figure}

In order to adapt our proposed attention model to image caption, the \emph{conditional global feature} should not only represent the weak visual feature descriptor as in the first conditional attention model, it  should also provide the context visual information for a language model.
Thus we adopt two layers of LSTM to extract visual and context features separately.
The first LSTM layer is aimed for our conditional visual attention to provide the \emph{conditional global feature} $CG_t$,
and the second LSTM layer works for the language model, specifically a bi-directional LSTM.
The adapted attention model for image caption is shown in Fig.~\ref{fig:cg2}.
At each time step, the attention LSTM requires the previous attention feature $g_{t-1}$,
the previous hidden state $h_{t - 1}^2$ of language LSTM, the previous hidden state $h_{t - 1}^1$ of attention LSTM,
and previous word vector $W_{t-1}$. Thus the $C{G_t}$ is produced given by Equation~\eqref{eq:CG_language}:
\begin{equation} \label{eq:CG_language}
C{G_t} = AttentionLSTM\left( {{g_{t - 1}},h_{t - 1}^1,h_{t - 1}^2,{W_{t - 1}}} \right)
\end{equation}
% Then in this sense, to calculate the \emph{conditional global feature} $CG_t$, The input vector to the attention LSTM not only provide the previous attention features and partially generated words of the previous step,
% but also provide the previous state of the language model, which is beneficial for the prediction of the next steps.

Then $C{G_t}$ is used to generate the attention feature $g_t$ through conditional attention submodule as discussed above.
For the incorporated language model, it requires the visual attention feature $g_t$, the \emph{conditional global feature} $C{G_t}$,
and the previous hidden state $h_{t - 1}^2$ of language LSTM, which provide not only the last visual attention, but also the context information of both visual and language.
Finally, $g_t$ and $h_t^2$ are concatenated to feed to the classifier to predict the conditional distribution over possible output at each step $t$, referring to Equation~\eqref{eq:prob_language}.
\begin{equation}\label{eq:prob_language}
p\left( {{y_t}|{g_{t-1},{h_{t-1}^2},{W_{t-1}},{h_{t-1}^1}}} \right) = f\left( {{g_t},{{h_{t}^2}}} \right)
\end{equation}
where \emph{f} is a classifier with a Softmax function that outputs the probability of $y_t$, $g_t$ is the attention feature vector at time \emph{t} and ${h_{t}^2}$ is the output of the language LSTM.

\subsection{Optimization}

%To train conditional attention model for the sequential visual task,
%given a target ground truth sequence $\left\{ {{y_1},{y_2}, \cdot  \cdot  \cdot ,{y_K}} \right\}$, where $K$ is the number of objects in the image
%or the length of the caption.
%we take the following training steps to optimize the model.
%
%Step1: the model receives an image and obtains \emph{L}, \emph{G} through a series of convolutional and nonlinear layers.
%The vector \emph{G} is used as the initial state ${h_0}$ of LSTM, the attention feature ${g_0}$ is generated by \emph{L} and \emph{G} alignment
%and used to predict the first object category label in the image.
%
%
%Step2: the model receives ${h_0}$ and ${g_0}$ as input of LSTM and predicts ${CG_1}$ as the first \emph{conditional global feature} in our model.
%The conditional attention submodule emits ${g_1}$ through compatible operation between \emph{L} and ${CG_1}$.
%The model updates its internal state ${h_1}$ according to its previous state ${h_0}$ and previous object's attention feature ${g_0}$.
%And the next step repeats the operation of the current step.

Optimizing our proposed conditional attention models is end-to-end, which allows the model to be trained directly with respect to a given task.
Back-propagation is exploited to train the neural network components by minimizing the smooth cross entropy loss function shown as Equation~\eqref{eq:xent}:
\begin{equation}\label{eq:xent}
J =  - \frac{1}{{N \times T}}\sum\limits_{i = 0}^{N -1} {\sum\limits_{t = 0}^{T - 1} {\sum\limits_{k = 0}^{K - 1} {{y_{i,t,k}}\log \left( {{p_{i,t,k}}} \right)} } }
\end{equation}
where \emph{N} is the number of samples, \emph{T} is the length of a sequence in a sequence task and $K$ is the number of categories.
${{y_{i,t,k}}}$ is the ground truth of class \emph{k} in step \emph{t} while ${{p_{i,t,k}}}$ is the predicted probability of class \emph{k} in step \emph{t}.

For the proposed attention model for image captioning, since a language model is incorporated, the training process is slightly different from the version of the multiple objects recognition task.
%The details of the model and the whole process are mentioned in Section \ref{subsec:3.3}, and will not be discussed here.
%A little attention is needed here, for fair comparison with recent work~\cite{xu2015show},
We adopt double stochastic attention loss as a regularization method follows by~\cite{xu2015show} to improving overall BLEU score:
\begin{equation}
{J_2} = \lambda \sum\limits_i^L {{{\left( {1 - \sum\limits_t^K {{\alpha _{ti}}} } \right)}^2}}
\end{equation}
where $\lambda$ is a constant, $L$ is the number of spatial locations of attention feature and $K$ is the length of the caption.
Thus the final loss function of training the attention model is:
\begin{equation}
{J} = - \frac{1}{{N \times T}}\sum\limits_{i = 0}^{N -1} {\sum\limits_{t = 0}^{T - 1} {\sum\limits_{k = 0}^{K - 1} {{y_{i,t,k}}\log \left( {{p_{i,t,k}}} \right)} } } + {J_2}
\end{equation}

% Numerical Experiments

\section{Experiments}
\label{sec:4}

%In the experiment, we show the performance of our conditional attention model, we first
%apply the proposed model to a multiple object
%recognition task using the multi-digit street view house number (SVHN) dataset~\cite{netzer2011reading}
%which contain bounding boxes for each digit.
%Next, for a more challenging task, we generate a 128$\times$128 SVHN dataset by enlarging the bounding box of each image
%like~\cite{ba2014multiple} where the task is to locate and identify the digits in this large map.
%Next, to reflect the generalization performance of our model on real scene data,
%We tested the performance of our model on the original SVHN dataset without bounding boxes.
%Finally, we evaluated

In the experiments, we first benchmarked the performance of our attention model for multiple objects recognition
on the SVHN dataset~\cite{netzer2011reading} with / without extra bounding box.
Next, we evaluated the our attention framework adapted for the image caption task on the MSCOCO 2014 captions dataset~\cite{lin2014microsoft}. 
\footnote{Our code for both multiple objects recognition and image caption has been released at \url{https://github.com/caoquanjie/ConditionalLearnToPayAttention}.}

\subsection{Experimental Setup}

\textbf{Attention model for multiple objects recognition}: The network structure configuration parameters in Fig.~\ref{fig:cg} are the same as~\cite{jetley2018learn} 
shown in Table~\ref{tab:att-cnn-network}. 
It has 16 layers: 15 convolutional layers and 1 fully connected layer. Different from the standard VGGNet~\cite{simonyan2014very}, the first two max-pooling layers of the VGGNet are moved 
after the additional  convolutional layers \emph{conv6} and \emph{conv7} respectively to make the estimated attention map have higher resolution.
The output of last layer is defined as the global feature \emph{G} and {the convolutional features \emph{conv4\_3} and \emph{conv5\_3} are defined as the local features ${L_1}$ and ${L_2}$.}
In all subsequent experiments, our model only with ${L_2}$ is referred as \emph{conditional-model-att1}, and the attention model with both ${L_1}$ and ${L_2}$ is referred as \emph{conditional-model-att2}.
In order to compare with our proposed models, we evaluated two versions of soft attention model~\cite{xu2015show}
as the baselines in the experiment.
One is \emph{standard-soft-attention}, saying that we use the standard VGGnet for soft attention according to its original implementation~\cite{xu2015show};
the other is \emph{modified-soft-attention}, in which the first two max pooling layers are removed from the VGGnet as ours.
Moreover, batch normalization and Dropout regularization are adopted to each convolutional layer, with dropout rate 0.3 for the first convolutional layer and 0.4 for the rest.
The implementation of LSTM closely follows the one used in~\cite{xu2015show},
the initial memory state and hidden state of the LSTM were predicted by global features through two fully connected layers mapping.
The size of the LSTM cell in our model was set to 512.	

\begin{table}[!htp]
\centering
\small
\caption{\label{tab:att-cnn-network}The CNN network of our attention model for multiple objects recognition.}
\setlength{\tabcolsep}{6.0mm}
{
\begin{tabular}{lclll}
 \toprule
 Layers&Filters&Stride&Depth \\
 \midrule
conv1\_1 &3 $\times$ 3 &1 $\times$ 1&64\\
\specialrule{0em}{-1pt}{-1pt}
conv1\_2 &3 $\times$ 3 &1 $\times$ 1&64\\
\specialrule{0em}{-1pt}{-1pt}
conv2\_1 &3 $\times$ 3 &1 $\times$ 1&128 \\
\specialrule{0em}{-1pt}{-1pt}
conv2\_2 &3 $\times$ 3 &1 $\times$ 1&128\\
\specialrule{0em}{-1pt}{-1pt}
conv3\_1 &3 $\times$ 3 &1 $\times$ 1&256\\
\specialrule{0em}{-1pt}{-1pt}
conv3\_2 &3 $\times$ 3 &1 $\times$ 1&256\\
\specialrule{0em}{-1pt}{-1pt}
conv3\_3 &3 $\times$ 3 &1 $\times$ 1&256\\
\specialrule{0em}{-1pt}{-1pt}
max\_pool &2 $\times$ 2 &2 $\times$ 2&256\\
\specialrule{0em}{-1pt}{-1pt}
conv4\_1 &3 $\times$ 3 &1 $\times$ 1&512\\
\specialrule{0em}{-1pt}{-1pt}
conv4\_2 &3 $\times$ 3 &1 $\times$ 1&512\\
\specialrule{0em}{-1pt}{-1pt}
conv4\_3 &3 $\times$ 3 &1 $\times$ 1&512\\
\specialrule{0em}{-1pt}{-1pt}
max\_pool &2 $\times$ 2 &2 $\times$ 2&512\\
\specialrule{0em}{-1pt}{-1pt}
conv5\_1 &3 $\times$ 3 &1 $\times$ 1&512\\
\specialrule{0em}{-1pt}{-1pt}
conv5\_2 &3 $\times$ 3 &1 $\times$ 1&512\\
\specialrule{0em}{-1pt}{-1pt}
conv5\_3 &3 $\times$ 3 &1 $\times$ 1&512\\
\specialrule{0em}{-1pt}{-1pt}
max\_pool &2 $\times$ 2 &2 $\times$ 2&512\\
\specialrule{0em}{-1pt}{-1pt}
conv6 &3 $\times$ 3 &1 $\times$ 1&512\\
\specialrule{0em}{-1pt}{-1pt}
max\_pool &2 $\times$ 2 &2 $\times$ 2&512\\
\specialrule{0em}{-1pt}{-1pt}
conv7 &3 $\times$ 3 &1 $\times$ 1&512\\
\specialrule{0em}{-1pt}{-1pt}
max\_pool &2 $\times$ 2 &2 $\times$ 2&512\\
\specialrule{0em}{-1pt}{-1pt}
fc & —— &—— &512\\
 \bottomrule
\end{tabular}
}
\end{table}

\textbf{Attention model for image caption}: We used the VGGnet pretrained on ImageNet as the convolutional layers in Fig.~\ref{fig:cg2}.
The configuration of the LSTM layer for visual attention is the same as our basic conditional attention model.
A bi-directional LSTM is exploited for the LSTM layer in this language model. The dimension of word embedding and the hidden state of LSTM were all set to 512.
In the testing period, we utilized beam search to select the best caption from some candidates, and the beam size was set to 3.
$\lambda$ in double stochastic attention was set to 1.

All models were trained using ADAM optimization algorithm~\cite{kingma2014adam} with mini-batch size of 32,
and the learning rate was set to 1e-4.
All our experiments were implemented by Tensorflow 1.9 and were trained on a workstation with NVIDIA 2080Ti GPU and 32Gb system RAM.

\subsection{Multiple Objects Recognition on SVHN}
\label{subsec:4.1}

The Street View House Numbers (SVHN) dataset~\cite{netzer2011reading} consists of three parts, the training set, the testing set, and the extra set, which has around 240k images in total. In the experiments, our model was trained by using the extra set and the testing set. In the first experiment on SVHN, the ground truth bounding boxes were used to generated the well conditioned dataset for multiple objects recognition. The dataset was preprocessed closely following \cite{goodfellow2013multi}, in which the digits were cropped from the street number images by using the bounding boxes, and the cropped digits were resized to $64 \times 64$. Then we randomly cropped $54 \times 54$ on the enlarged digits to augment our training dataset.
Since there are at most 5 digits in one street number image,
we used a dummy label '10' as a placeholder to guarentee the label length of each image to be 5.

% In this experiment, the evaluation of our model performance is shown in the Table \ref{tab:err-rates}.
% Results in the table demonstrate that our model obtains better performance than existing methods.
% But the local features of layer 13 in the CNN pipeline appears to focus on the central object which
% results in incomplete features for classification.
% We therefore train a second conditional attention model: \emph{conditional-model-att2}.
% Adding one layer of local features with another scales in our model works very well in practice,
% and obtain a state-of-the-art accuracy of 97.11$\%$, outperform DRAM~\cite{ba2014multiple},~\cite{jaderberg2015spatial}
% and~\cite{xu2015show} on the SVHN dataset, because different convolution layers contain features of multiple scales, which make the focus on object more diverse. The recognition accuracy of our models is shown in Fig \ref{fig:4}. %\ref{fig:2}.
\begin{table}[!htb]
\centering
 \caption{\label{tab:err-rates}Whole sequence recognition accuracy on SVHN dataset with trained models.}
\setlength{\tabcolsep}{2.5mm}{
 \begin{tabular}{lcl}
  \toprule
  Model&Test acc \\
  \midrule
 11-layer-CNN (~\cite{goodfellow2013multi}) &96.04$\%$\\
 \specialrule{0em}{-1pt}{-1pt}
 10-layer-CNN &95.89$\%$ \\
 \specialrule{0em}{-1pt}{-1pt}
 DRAM(~\cite{ba2014multiple})&94.9$\%$\\
 \specialrule{0em}{-1pt}{-1pt}
 ST-CNN Single (~\cite{jaderberg2015spatial})&96.30$\%$\\
 \specialrule{0em}{-1pt}{-1pt}
 ST-CNN Multi (~\cite{jaderberg2015spatial})&96.40$\%$\\
 \specialrule{0em}{-1pt}{-1pt}
 standard-soft-attention (~\cite{xu2015show}) &96.47$\%$\\
 \specialrule{0em}{-1pt}{-1pt}
 modified-soft-attention &96.08$\%$\\
 \specialrule{0em}{-1pt}{-1pt}
 conditional-model-att1 (ours)   &96.98$\%$\\
 \specialrule{0em}{-1pt}{-1pt}
 \textbf{conditional-model-att2} (ours) &\textbf{97.15$\%$}\\
  \bottomrule
 \end{tabular}}
\end{table}

\begin{figure}[!tb]
\centering
\includegraphics[width=0.88\textwidth]{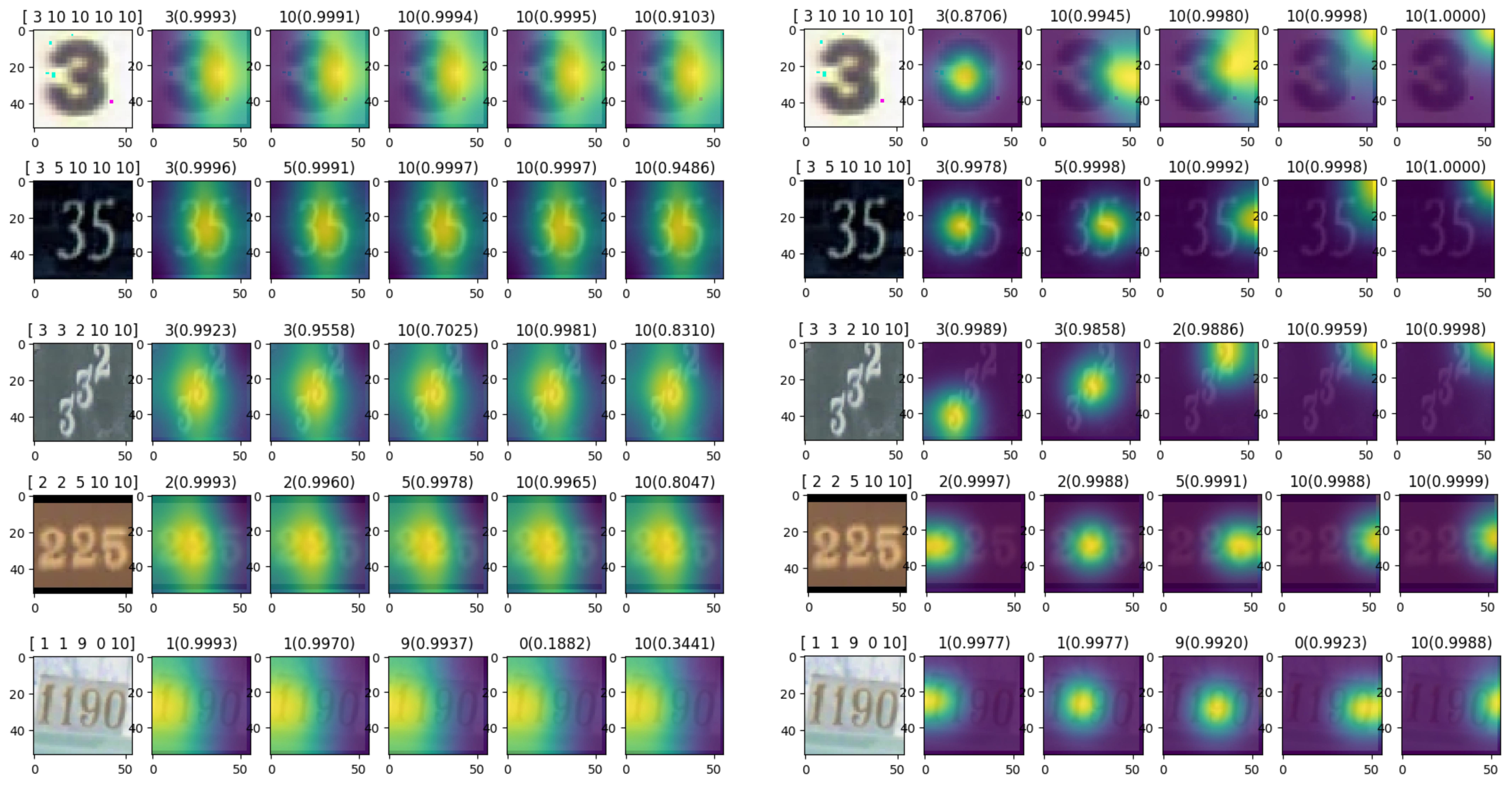}
\caption{Attention maps from \emph{standard-soft-attention}~\cite{xu2015show}(\textbf{left}) and \emph{conditional-model-att1}(\textbf{right}) trained on SVHN dataset with bounding box. Above the images are the probabilities of the label prediction. Our model learns to focus on the central part of each object.}
\label{fig:3}
\end{figure}

Table~\ref{tab:err-rates} demonstrates the performance of our model compared with state of the arts in this experiment. It can be seen that both \emph{conditional-model-att1} and \emph{conditional-model-att2} outperform other methods.
Compared with the existing methods, at each step, the obtained attention feature helps to improve the performance of our model by focusing on the object of interest while suppressing the background regions.
%In addition, since our model is end-to-end and differentiable, unlike DRAM~\cite{ba2014multiple}, it does not need to be trained by reinforcement learning technique.
Fig.~\ref{fig:3} visualizes the differences between the generated attention map by our proposed attention model and the soft attention model~\cite{xu2015show}.
It is interesting that our model can focus more accurately than the soft attention model~\cite{xu2015show} from left to right on each object in the image.
In addition, the probability of the predicted label is shown on the top of each image. Note that for the third very skewed image ``332'' in Fig.~\ref{fig:3}, our model correctly predicts all digits, while the soft attention model wrongly predicts the last digit `2', because our attention model directly computes the attention feature through the compatibility function between local features and the \emph{conditional global feature} at each time step, rather than relies on a simple attention network.
For the two versions of our model, \emph{conditional-model-att2} performs better than \emph{conditional-model-att1} since it exploits multi-scale attentions, however, \emph{conditional-model-att2} requires more training time.

\subsection{Weakly Supervised Multiple Objects Recognition}
\label{subsec:4.2}

\begin{figure}[!htb]
\centering
\includegraphics[width=0.88\textwidth]{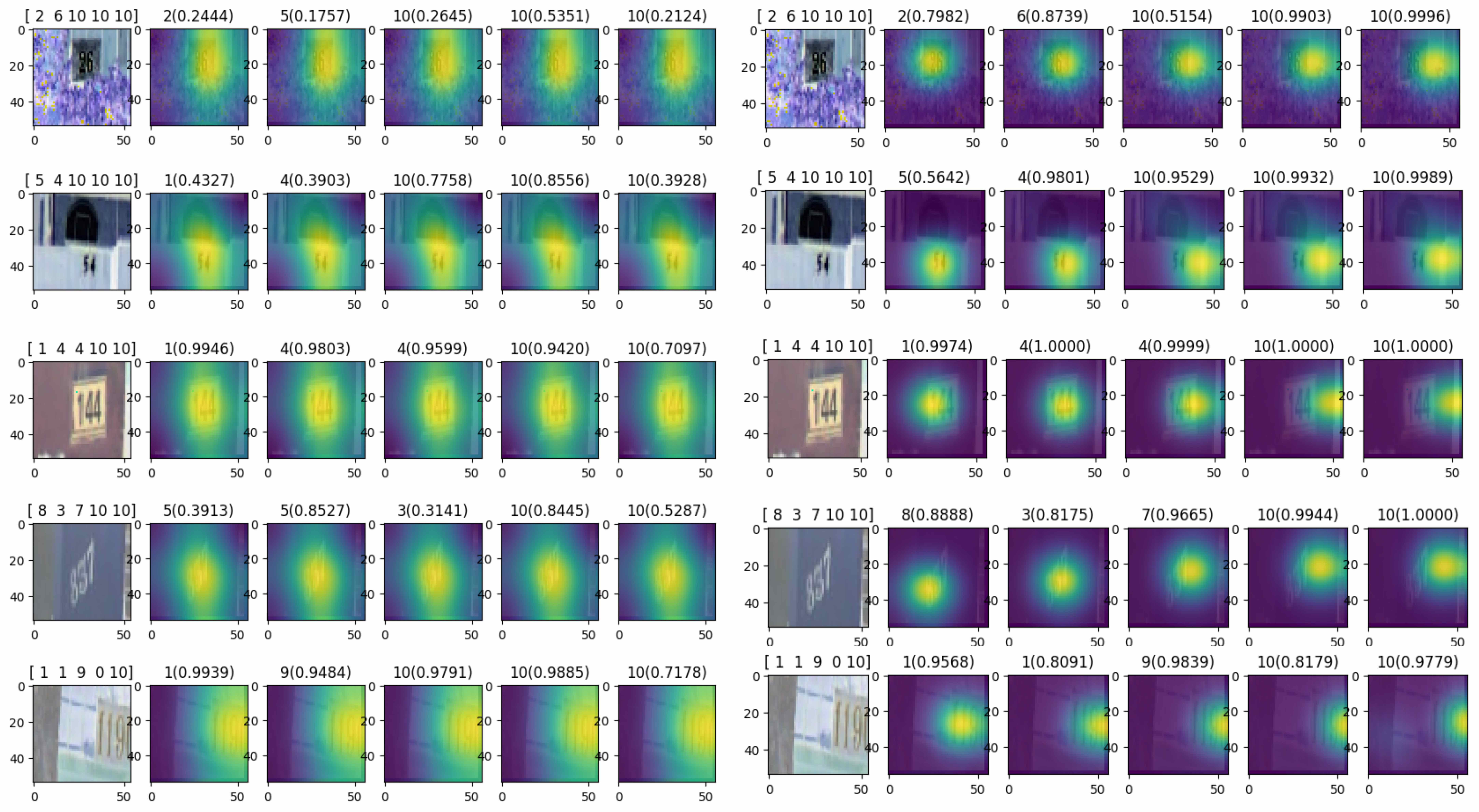}
\caption{Attention maps from \emph{standard-soft-attention}~\cite{xu2015show}(\textbf{left}) and \emph{conditional-model-att1}(\textbf{right}) trained on SVHN dataset without bounding box. Our model learns to focus on the central part of each object}
\label{fig:5}
\end{figure}

To further test the robustness of our model, we designed a challenging weakly labeled multiple objects recognition task in which no ground truth bounding boxes were used to crop the digits. Instead, the original real street number images are directly used for training and testing. The first column in Fig.~\ref{fig:5} shows the original street number images, say weakly labeled images in this experiment. We used the same data augmentation technique as previous experiments, saying that the original SVHN images were resized to $64 \times 64$ pixels and randomly cropped to $54 \times 54$. It can seen from the Table \ref{tab:err-rates3}, although the overall accuracy of all models is not too high, by comparison, our attention models especially \emph{conditional-model-att2} achieves $2.84\%$, $5.56\%$ and $9.6\%$ improvements over \emph{modified-soft-attention}, \emph{standard-soft-attention}~\cite{xu2015show} and \emph{11-layer-CNN}~\cite{goodfellow2013multi} respectively for this task.	

% In order to reflect the generalization performance of our model, we compared the experimental results on a more difficult task.
% SVHN dataset is composed of the real number of street buildings. Therefore, We scaled the original SVHN data to 64$\times$64 pixels and randomly crop to 54$\times$54, then directly recognize the original image containing the building background. As you can see from the Table \ref{tab:err-rates3}, although the overall accuracy of all models is not too high, by comparison, our attention models especially \emph{conditional-model-att2} achieves a 2.84$\%$, 5.56$\%$ and 9.6$\%$ improvement over \emph{modified-soft-attention}, \emph{standard-soft-attention}~\cite{xu2015show} and \emph{11-layer-CNN}~\cite{goodfellow2013multi} for this task.
\begin{table}[htbp]
\centering
 \caption{\label{tab:err-rates3}Weakly supervised multiple recognition accuracy on SVHN dataset.}
\setlength{\tabcolsep}{2.5mm}{
 \begin{tabular}{lcl}
  \toprule
  Model&Test acc \\
  \midrule
 11-layer-CNN (~\cite{goodfellow2013multi}) &70.58$\%$\\
 \specialrule{0em}{-1pt}{-1pt}
 standard-soft-attention (~\cite{xu2015show}) &74.89$\%$\\
 \specialrule{0em}{-1pt}{-1pt}
 modified-soft-attention &77.61$\%$\\
 \specialrule{0em}{-1pt}{-1pt}
 conditional-model-att1 (ours)&79.04$\%$\\
 \specialrule{0em}{-1pt}{-1pt}
 \textbf{conditional-model-att2} (ours) &\textbf{80.45$\%$}\\

  \bottomrule
 \end{tabular}}
\end{table}

Fig.~\ref{fig:5} visualizes the attention maps generated by our proposed attention model and the soft attention model~\cite{xu2015show}. It clearly shows that our model correctly highlights the street numbers at each step despite the very noisy and cluttered large background. However, the soft attention model can only coarsely identify the regions of street numbers. For example, for the extremely noisy street numbers image ``26'' at the first row in Fig.~\ref{fig:5}, our attention model can accurately predict each number with high confidence, say 0.7982 for `2' and 0.8739 for `6'; but the soft attention model wrongly recognizes `6' as `5' and the overall prediction probabilities are low, say 0.2444 for `2' and 0.1757 for `5'.

Note that this task is very	different from the classical object detection which requires the accurate bounding boxes for training. However, for this weakly supervised recognition task, only the sequential labels can be exploited. Experiment shows that our model can always nicely identify the interested objects. Thus we hypothesize that our model can be used to automatically crop the regions of interest (ROI) for collecting the training dataset of object detection by simply sequentially memorizing the interested objects, which would greatly reduce the burdening manual annotation work. We put this investigation in our future work. %Thus, in some sense, this experiment demonstrates that our model can help people to identify the 	

%Fig. \ref{fig:5} visualizes the attention maps on weak SVHN dataset. soft attention models from~\cite{xu2015show} are easy to identify digits errors while our models nicely predict the correct labels. Even if both models predict the correct label, the probabilities of the label predictions of our attention model are higher.

\subsection{Evaluation for Image Caption }

% \newpage
In our last experiment, we designed another \emph{conditional global feature} based on a language model. 
Note that the main purpose of this evaluation is to demonstrate the generality of our conditional attention framework 
on the image caption task, and thus we do not expect it could outperform the state of the arts methods optimized for image captioning, 
such as the bottom-up approach~\cite{anderson2018bottom}. However, we show that our attention model does excel several popular models 
based on the VGGNet architecture as ours on the image caption task. 
%since we only consider to 
We used the MSCOCO 2014 captions dataset~\cite{lin2014microsoft} to evaluate the our attention model,
%The MSCOCO is the largest image captioning dataset, containing 82783, 40504 and 40775 images for training, validation and test respectively,
%and each image has 5 human annotated captions. To facilitate parameter tuning and test model performance,
and used the same split as~\cite{karpathy2015deep}. % to facilitate parameter tuning and test model performance.
%, which contains 5000 images respectively for validation and testing.
To deal with various length	of captions, we truncated each caption to 20 words. % for MSCOCO.
The evaluation criteria  are the standard automatic evaluation metrics, namely BLEU~\cite{papineni2002bleu}, METEOR~\cite{denkowski2014meteor}, and CIDEr~\cite{vedantam2015cider}.

\begin{table*}[!htb]
\centering
\small
 \caption{\label{tab:caption1}The model performance on MSCOCO test splits.
 Our attention model outperforms many existing models
 especially in the score of B-2 to B-4.}
\setlength{\tabcolsep}{5mm}{
 \begin{tabular}{lclllcc}
  \toprule
  Method&B-1&B-2&B-3&B-4&METEOR&CIDEr \\
  \midrule
 Deep VS~\cite{karpathy2015deep}&62.5&45.0&32.1&23.0&19.5&-\\
 \specialrule{0em}{-1pt}{-1pt}
 m-RNN~\cite{mao2014deep}&67.0&49.0&35.0&25.0&-&-\\
 \specialrule{0em}{-1pt}{-1pt}
 Google NIC~\cite{vinyals2015show}&66.6&46.1&32.9&24.6&-&-\\
 \specialrule{0em}{-1pt}{-1pt}
 Soft-Attention~\cite{xu2015show}&70.7&49.2&34.4&24.3&23.9&-\\
 \specialrule{0em}{-1pt}{-1pt}
 Hard-Attention~\cite{xu2015show}&71.8&50.4&35.7&25.0&23.04&-\\
 \specialrule{0em}{-1pt}{-1pt}
 SCA-CNN-VGG~\cite{chen2017sca}&70.5&53.3&39.7&29.8&24.2&89.7\\
 \specialrule{0em}{-1pt}{-1pt}
 SCA-CNN-ResNet~\cite{chen2017sca}&71.9&54.8&41.1&31.1&25.0&95.2\\
 \specialrule{0em}{-1pt}{-1pt}
 \textbf{Our attention model}&70.9&54.1&40.5&30.3&23.9&89.5\\

  \bottomrule
 \end{tabular}}
\end{table*}

Results from Table \ref{tab:caption1}, we can see that our attention model adapted for image captioning outperforms the other models in most cases.
Especially when compared with soft attention~\cite{xu2015show}, our model sightly outperforms it on the score of B-1 though, it
achieves large improvement over soft attention on the scores of B-2, B-3, and B-4.
%These improvements demonstrate that the attention feature computed by the compatibility function between local
%features and the \emph{conditional global feature} does effectively focus on important areas of an image, and thus .
Compared with the recent proposed SCA-CNN models~\cite{chen2017sca}, given the same pretrained VGGnet, our model outperforms SCA-CNN-VGG on all BLEU scores,
even though SCA-CNN-VGG incorporates spatial and channel-wise attentions and exploits multi-layer attentions~\cite{chen2017sca}. 
For a more powerful CNN architecture,
such as ResNet, we expect our model could benefit more to produce more accurate visual attention, which in turn to generate better caption results, just as
SCA-CNN-ResNet. We put it as our future investigation.
% However, our results are not good enough to compare with SCA-CNN-ResNet. The reasons come from two sides: on one hand,
% we only use a pretrained VGGnet for image caption rather than more powerful CNN models such as ResNet exploited in SCA-CNN-ResNet; on the other
% hand, SCA-CNN-ResNet incorporates spatial and channel-wise attentions in CNN and considers multi-layer attentions,
% which can significantly improve the performance.	

\begin{figure*}[!htb]
\centering
\includegraphics[width=.8\textwidth]{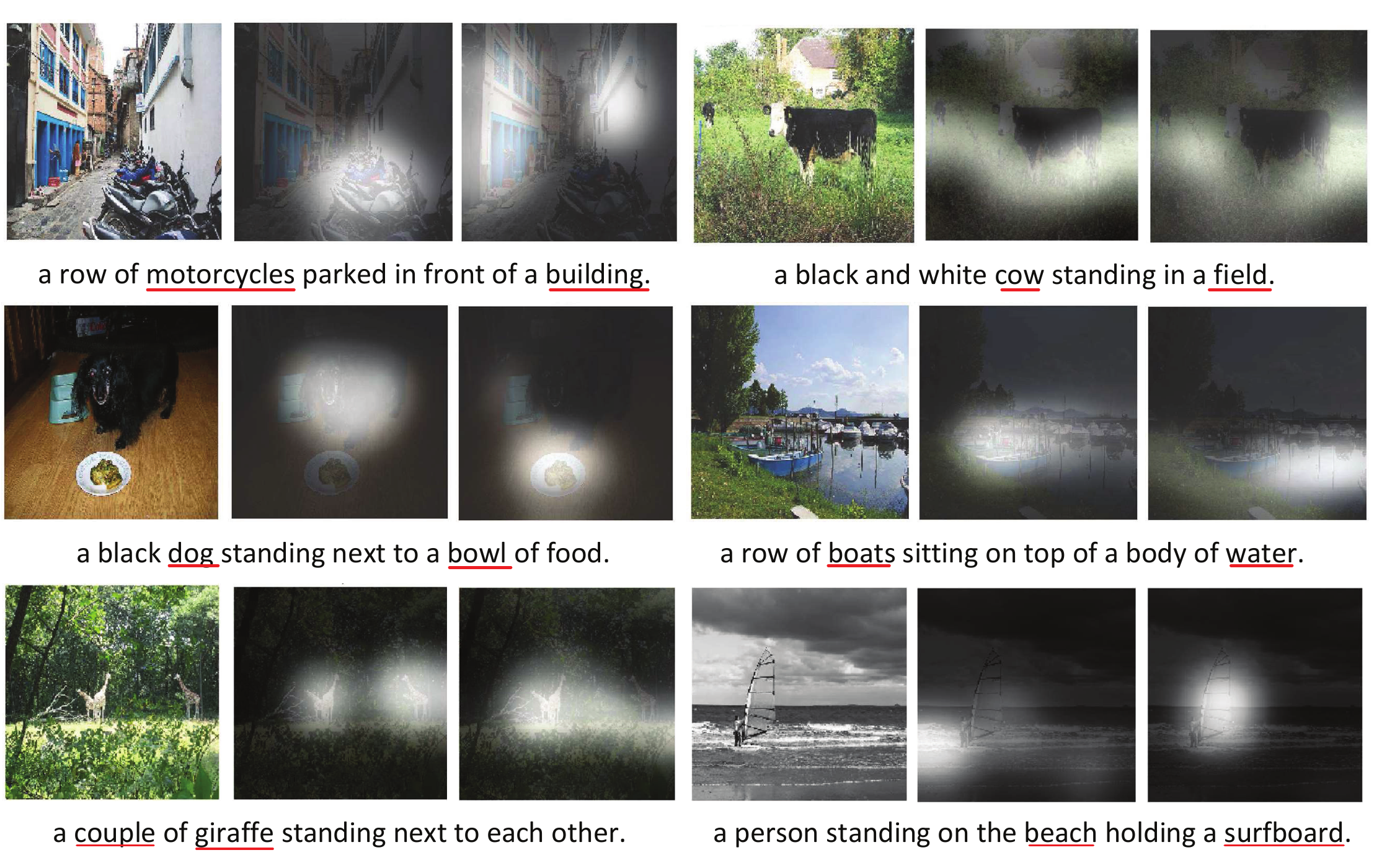}
\caption{Examples of generated captions and corresponding visual attention maps on MSCOCO with proposed attention model adapted to image caption. These three rows show attended regions is consistent with the underlined words. Detailed captions and visualization of attention maps are included in the Appendix.}
\label{fig:caption}
\end{figure*}

Fig.~\ref{fig:caption} shows the visualization results of our attention model on the image caption task.
Our attention model generates meaningful captions for images and the computed
attention maps do highlight the relevant parts corresponding to the caption words in most cases.
%To compare with the soft attention model~\cite{xu2015show},
%However, in some cases, the robustness of the soft attention model from~\cite{xu2015show} is not strong enough.
Note that the two images of the last row in Fig.~\ref{fig:caption} are also exhibited in~\cite{xu2015show} as the failure cases. For the ``giraffe'' at the left side, our attention model can accurately recognize it as ``giraffe''. Moreover, it is interesting that the word ``couple'' does correspond to two giraffes, which indicates that our model can learn the basic counting capability.

\section{Conclusions}
\label{sec:5}

In this paper, we proposed a new conditional attention framework to tackle sequential visual tasks, such as multiple objects recognition and image captioning.
By introducing the novel {\it conditional global feature},  the attention feature for each object can be produced by measuring how the local convolutional feature aligns to the {\it conditional global feature}.
Moreover, the proposed conditional attention model has demonstrated its generality  on various sequential visual tasks by designing the {\it conditional global feature}, which has achieved the best performance on the SVHN benchmark with / without extra bounding box,
and generated better scores on the image captioning task than the popular soft attention model.
%For multiple objects recognition and image caption, the {\it conditional global feature} can be generated either by a simple one-layer recurrent neural network or a language model.  Since our proposed model ,
Thus we believe that the new conditional attention model could also achieve promising performance for other sequential visual tasks, such as VQA. We put it as our future work.

%\section*{References}
\bibliographystyle{unsrt}
\bibliography{attention_arxiv}

\newpage

\begin{appendices}
\section{Appendices}

\subsection{Visualization of Image Captioning}
%Visualization results of the image caption task, attention focus on the regions of the image and predicts the corresponding words.
\begin{figure*}[htbp]
\centering
\includegraphics[width=.88\textwidth]{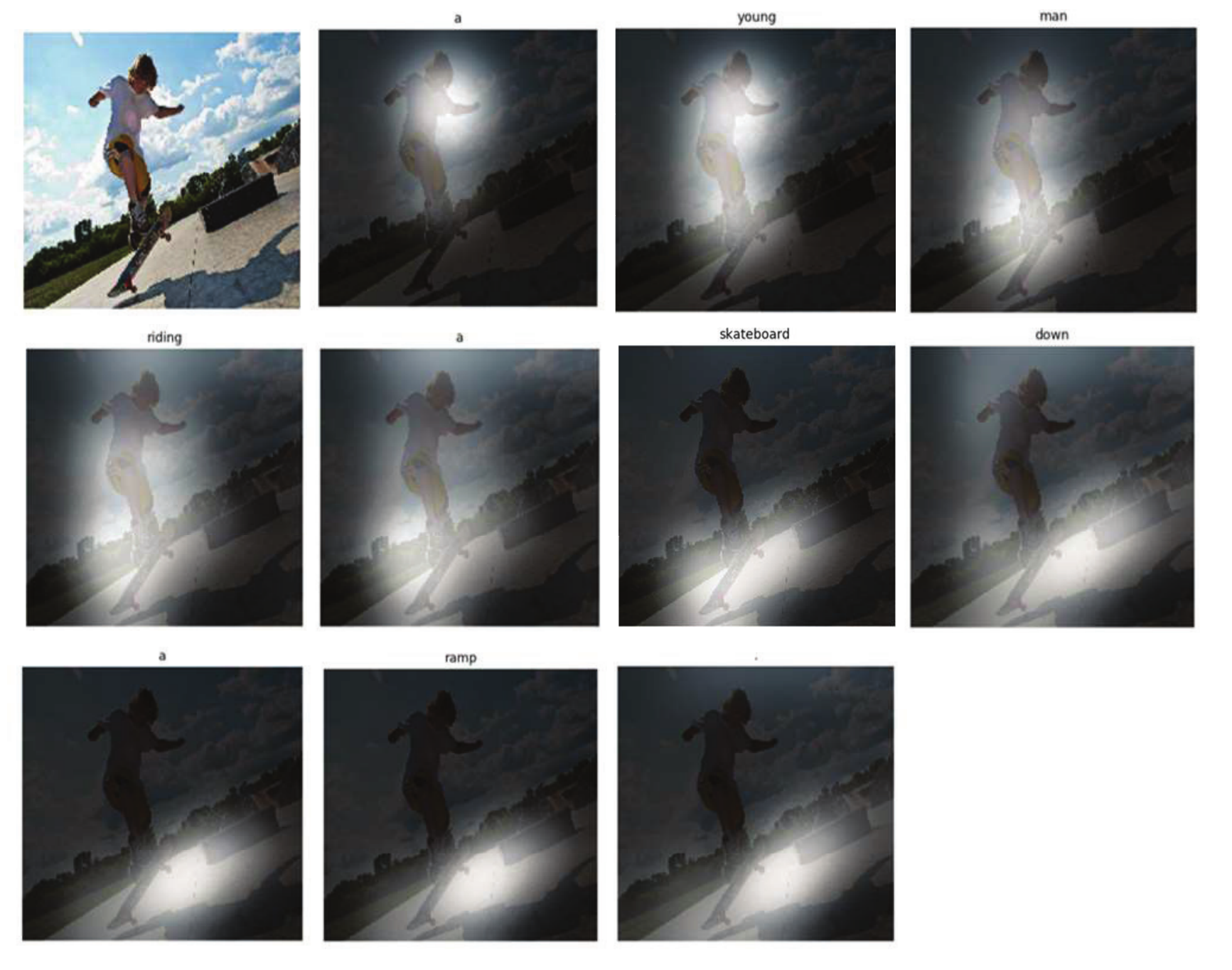}
\caption{a young man riding a skateboard down a ramp.}
\label{fig:caption1}
\end{figure*}

\begin{figure*}[htbp]
\centering
\includegraphics[width=.88\textwidth]{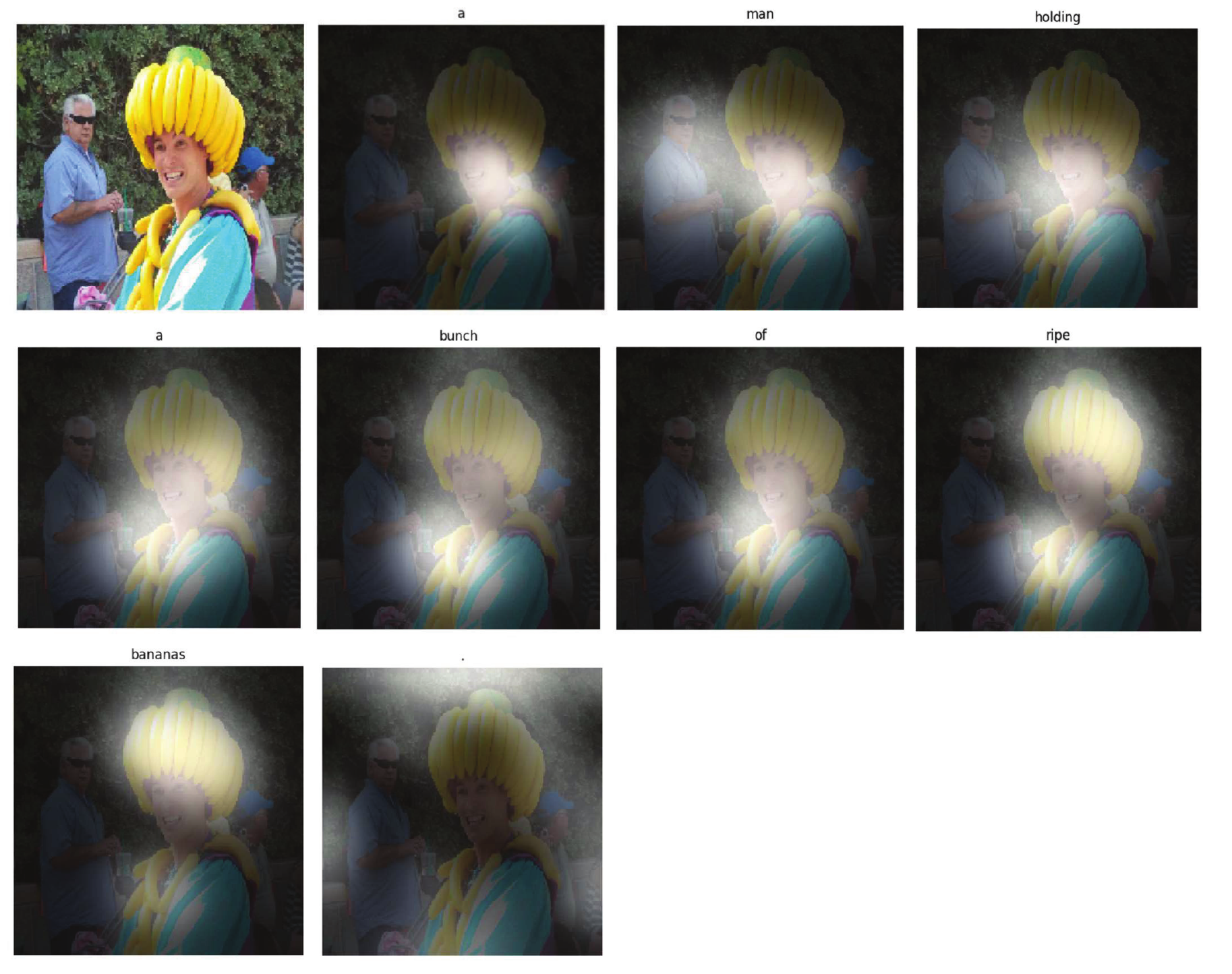}
\caption{a man holding a bunch of ripe bananas.}
\label{fig:caption2}
\end{figure*}

\begin{figure*}[htbp]
\centering
\includegraphics[width=.88\textwidth]{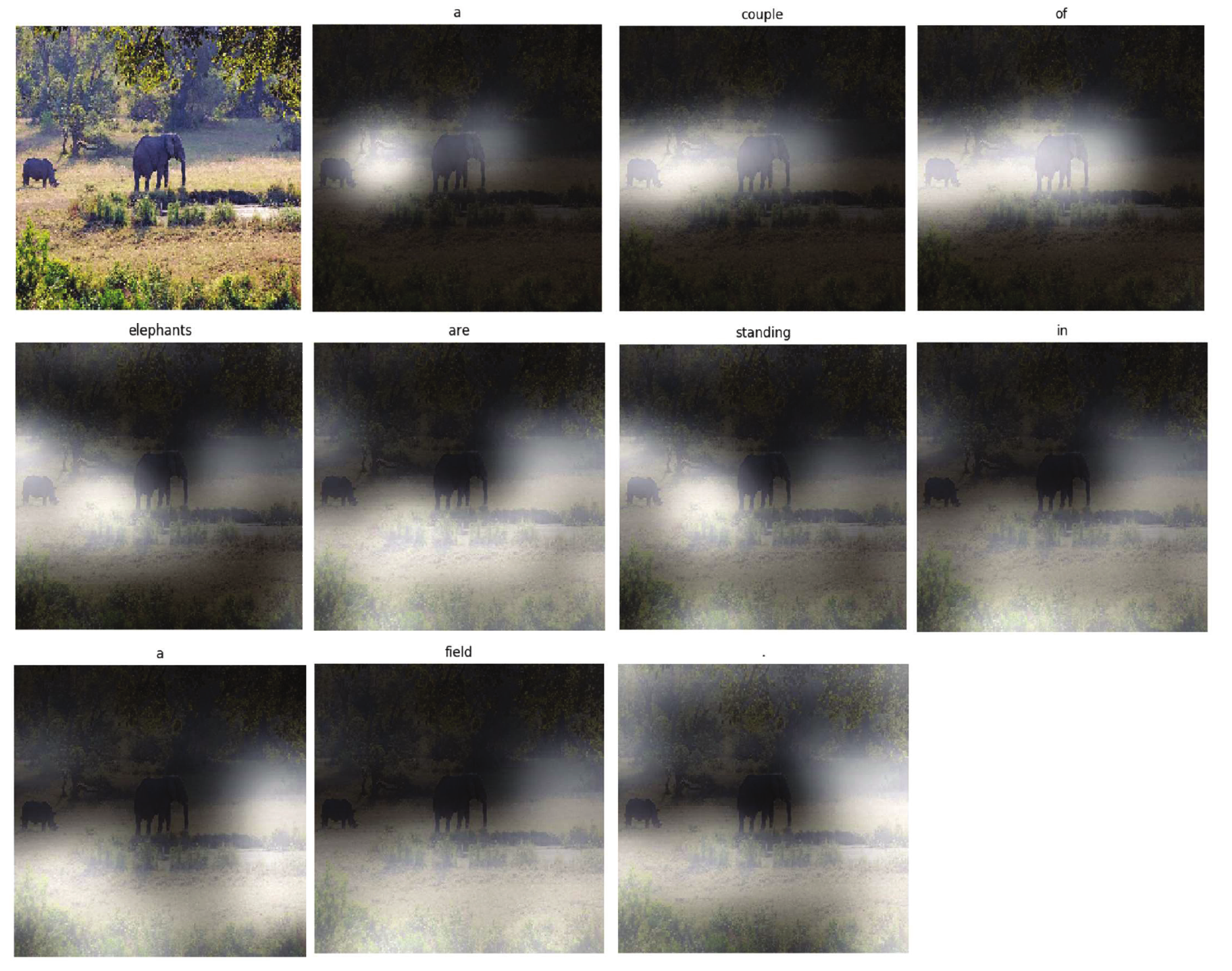}
\caption{a couple of elephants are standing in a field.}
\label{fig:caption3}
\end{figure*}

\begin{figure*}[htbp]
\centering
\includegraphics[width=.88\textwidth]{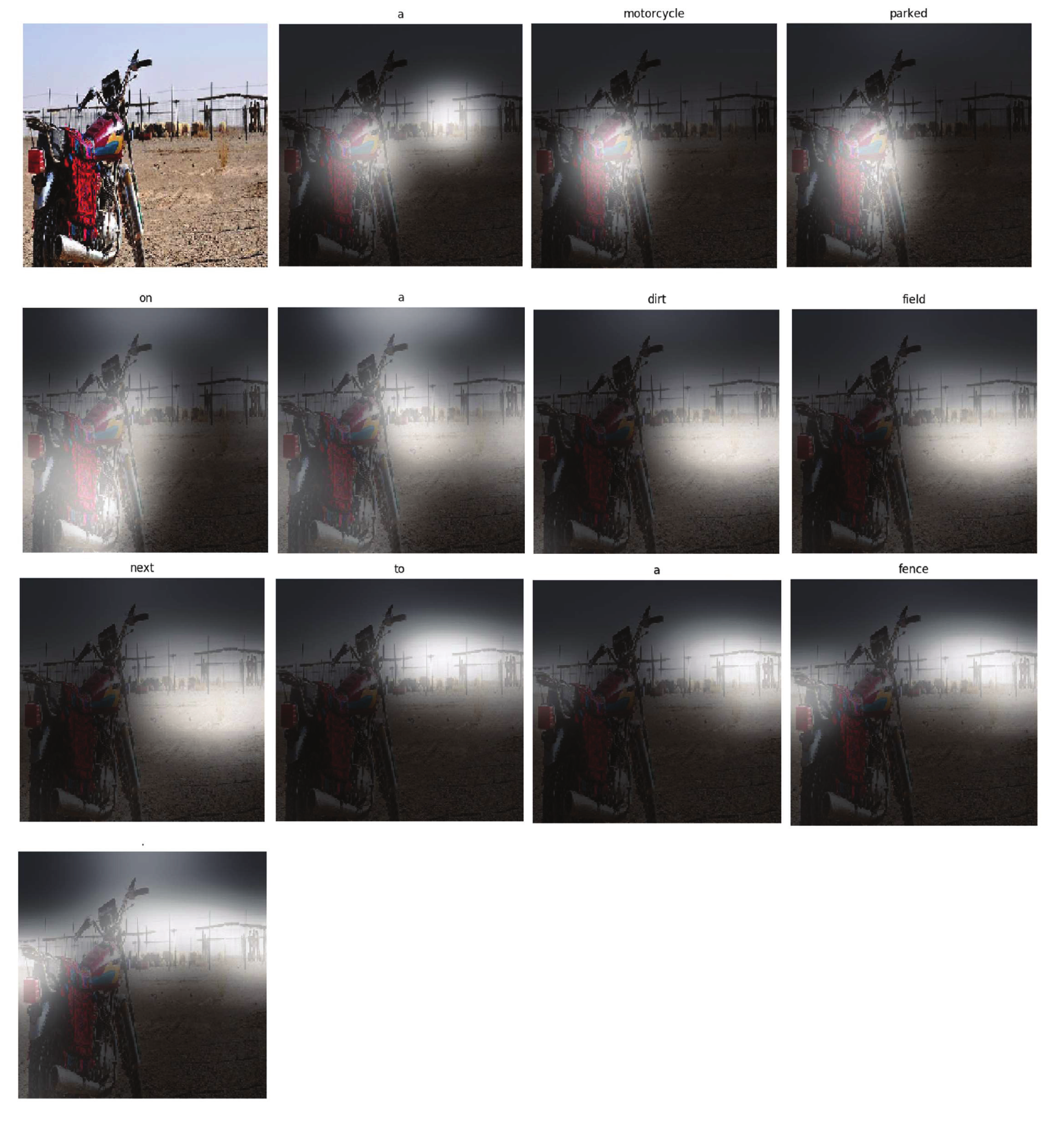}
\caption{a motorcycle parked on a dirt field next to a fence.}
\label{fig:caption4}
\end{figure*}

% \begin{figure*}[htbp]
% \centering
% \includegraphics[width=.88\textwidth]{caption5}
% \caption{a row of boats sitting on top of a body of water.}
% \label{fig:caption5}
% \end{figure*}

% \begin{figure*}[htbp]
% \centering
% \includegraphics[width=.88\textwidth]{caption6}
% \caption{a row of motorcycles parked in front of a building.}
% \label{fig:caption6}
% \end{figure*}

% \begin{figure*}[htbp]
% \centering
% \includegraphics[width=.88\textwidth]{caption7}
% \caption{a black and white cow standing in a field.}
% \label{fig:caption7}
% \end{figure*}

% \begin{figure*}[htbp]
% \centering
% \includegraphics[width=.88\textwidth]{caption8}
% \caption{a black dog standing next to a bowl of food.}
% \label{fig:caption8}
% \end{figure*}

% \begin{figure*}[htbp]
% \centering
% \includegraphics[width=.88\textwidth]{caption9}
% \caption{a couple of giraffe standing next to each other.}
% \label{fig:caption9}
% \end{figure*}

% \begin{figure*}[htbp]
% \centering
% \includegraphics[width=.88\textwidth]{caption10}
% \caption{a person standing on the beach holding a surfboard.}
% \label{fig:caption10}
% \end{figure*}

\end{appendices}

\end{document}